\documentclass[letterpaper]{article}

\usepackage{cite, graphicx, amsmath, amsfonts, amssymb, array, latexsym, url, algorithm, algorithmic}
\usepackage{multirow, booktabs}

\newcommand{\xx}{{\bf x}}
\newcommand{\zz}{{\bf z}}
\newcommand{\ls}{\varphi(\xx)}

\begin{document}

\title{Image Segmentation Using Weak Shape Priors}

\author{Robert~Sheng~Xu,~Oleg~Michailovich, and~Magdy~Salama
\thanks{This research was supported by a Discovery grant from NSERC -- The Natural Sciences and Engineering Research Council of Canada. Information on various NSERC activities and programs can be obtained from {\tt http://www.nserc.ca}.}%
\thanks{R. Xu, O. Michailovich, and M. Salama are with the School of Electrical and Computer Engineering, University of Waterloo, Canada N2L 3G1 (phone: 519-888-4567; e-mails: {rsxu, olegm, M.Salama}@uwaterloo.ca).}}

\maketitle

\begin{abstract}
The problem	 of image segmentation is known to become particularly challenging in the case of partial occlusion of the object(s) of interest, background clutter, and the presence of strong noise. In the case when the segmentation is performed by means of active contours, a typical way to overcome the above difficulties is to impose an {\it a priori} model on the shape of the contours -- a model which can be either analytical or probabilistic in its nature. Effectively, the model represents some shape constraints, which are intended to regularize the segmenting contour in the case when imagery data alone fails to provide sufficient information for determination of its optimal configuration. In practice, the shape models are typically learned based on training sets of examples of the object(s) of interest. In such cases, the goodness of modeling depend on the size of the training set, with more examples leading to more reliable and accurate models. Unfortunately, when the number of training samples is relatively small, the resulting model can be inadequate, thereby tending to bias the segmenting contour towards a suboptimal solution. To overcome this deficiency, the present paper introduces a novel approach to modeling of shape priors. Specifically, in the proposed method, an active contour is constrained to converge to a configuration at which its geometric parameters attain their empirical probability densities closely matching the corresponding model densities that are learned based on training samples. It is shown through numerical experiments that the proposed shape modeling can be regarded as ``weak" in the sense that it minimally influences the segmentation, which is allowed to be dominated by data-related forces. On the other hand, the priors provide sufficient constraints to regularize the convergence of segmentation, while requiring substantially smaller training sets to yield less biased results as compared to the case of PCA-based regularization methods. The main advantages of the proposed technique over some existing alternatives is demonstrated in a series of experiments.
\end{abstract}



\section{Introduction}\label{Intro}
Image segmentation is know to be a problem of fundamental importance in numerous applications of computer vision and image processing \cite{Osher03}, some standard examples of which include (yet not limited to) medical imaging \cite{McInerney96, Tsai03}, surveillance \cite{Stringa00, Haritaoglu00}, robotics \cite{Nair98}, and control \cite{Todorovic04}. In all these applications, image segmentation is employed to partition a data image (or a sequence thereof) into a number of its fragments which are associated with different {\em classes}, normally an object and its background. In 2-D imaging, such a partition can be performed using deformable curves (also known as {\em active contours}), whose optimal configuration maximizes a measure of statistical dissimilarity between the segmentation classes. Thus, at some fundamental level, the segmentation is based upon the fact that, in a properly defined domain of image features, the object and its background may have distinct statistical characteristics. Unfortunately, the validity of this assumption is compromised in the presence of strong noises and partial occlusions, which necessitates the development of robust segmentation tools.

In the past, a variety of classical algorithms \cite{Kass87, Mumford89, Morel95, Caselles97, ChanVese01} were proposed to segment objects of interest based on imagery data alone (see also \cite{Aubert03} and references therein for a summary of such techniques). Through employing edge-related and region-based features of data images, the above methods are capable of providing reliable segmentation results on conditions of moderate noises and unoccluded objects. Unfortunately, the methods are known to be prone to erroneous segmentation in practical scenarios, in which some parts of the object of interest appear to be occluded, missing, or corrupted by strong noises. In these situations, a standard remedy is to enhance the image segmentation through the introduction of prior shape knowledge.

A variety of segmentation algorithms incorporating shape priors have been proposed in the literature~\cite{Leventon00, Chen02, Paragios02, Cremers02, Tsai03, Chan03, Freedman05, Bresson06}. Most of these algorithms poses the segmentation task as a combination of two competing optimization problems: whilst the first problem maximizes the likelihood of contour's configuration based on image-related information, the second problem minimizes the extent to which the contour violates the shape constraints imposed by the prior model. Thus, for example, the prior model in \cite{Chen02} is defined by the average shape of a set of training segmentations. Subsequently, the active contour is constrained to deviate the least from this average. A similar approach is exploited in \cite{Chan03}, with a different cost function used to assess the above deviation. While useful in the situations when the shape of the active contour needs to be strongly constrained (as it would be the case with occlusions), the methods of \cite{Chen02, Chan03} can result in useless segmentation when the actual shape of the object happens to deviate considerably from its sample average.

In \cite{Leventon00}, a parametric shape model is derived by performing principle component analysis (PCA) on a training set of level-set functions \cite{Sethian99}. Subsequently, the level-set function representing the actual active contour is modeled as a linear combination of the principal components (also known as eigenshapes \cite{Tsai_med}) offset by the algebraic mean of the training functions. A slightly different formulation of the same idea is presented in \cite{Tsai03}, which uses a PCA-based shape modeling in combination with the region-based active contours of \cite{ChanVese01}. As will be demonstrated later in this paper, one of the main drawbacks of the PCA-based shape modeling lies in the dependency of its validness on the size of the training sets in use. In particular, small training sets lack the ability to represent all possible appearances of the desired object, and hence the resulting PCA model could introduce a significant bias in the estimated shape. Moreover, since it is based on Euclidean-type metrics, PCA could be easily affected by misrepresentative training samples (outliers), whose negative influence will be particularly destructive in the case of small training sets. Finally, the PCA-based methods \cite{Leventon00, Tsai03, Tsai_med} require the use of an alignment procedure which must be performed at each iteration of the segmentation process. This alignment step is posed as an additional minimization problem, which substantially increases the overall computational complexity of segmentation.

In a more recent work reported in \cite{Freedman05}, a segmentation method based on the concept of distribution tracking has been proposed. The method of \cite{Freedman05} is based on learning {\em both} the shape and appearance models, and it is designed in a way that eliminates the need to compute pixel-wise correspondences for alignment purposes. Moreover, \cite{Freedman05} alleviates the problem of high dimensionality associated with level set functions via performing PCA on a smaller set of control points representing the training shapes. Although the above algorithm appears to address some of the shortcomings of previous techniques, the reliability of its shape model still depends on the size of the training sets in use. Therefore, the associated disadvantages remain.

The method of \cite{Freedman05} is based on the earlier results of \cite{Freedman04}, in which the concept of segmentation via distribution tracking was first described. In particular, the idea in \cite{Freedman04} is to learn the probability distribution of a photometric parameter (or a set thereof) of the object class, followed by propagating the active contour towards a configuration at which the image region encompassed by the contour has the same parameter distributed similarly to the learned model. In this regard, the present work extends the idea of \cite{Freedman04} to tracking of {\em morphological} parameters/features. Specifically, given training data, the probability densities of a set of morphological {\em and} photometric parameters are estimated first. Subsequently, the active contour is propagated towards a configuration which results in a ``match" between both the photometric and morphological distributions. The latter can be seen as playing the role of a ``weak" shape prior\footnote{The notion of ``weakness" is introduced to stress the fact that many shapes may have an identical distribution of their geometric parameters, such as curvature.}, which nevertheless is informative enough to provide an effective regularization force. In particular, in our experimental study, we will show that the ``weak" model outperforms the PCA-based shape representation for the case of small training sets. Moreover, an additional advantage of the ``weak" model stems from the fact that the morphological parameter(s) can be chosen to be invariant under a class of geometrical transformations (such as Euclidean or affine), which effectively eliminates the need for intermediate alignment routines.

The rest of the paper is organized as follows. In Section II, a method for image segmentation via tracking both texture-related and morphological features is detailed. Some essential numerical details are provided in Section III. Section IV presents a series of experimental results, which demonstrate some principal advantages of the proposed technique. Finally, Section V summarizes the paper with a discussion, conclusions, and an outline of our future research.

\section{Tracking of distributions}

\subsection{Tracking of photometric features}
Let $I$ be a scalar-valued image defined over an open subset $\Omega \in \mathbb{R}^2$. To make the discussion general, the image $I$ can be transformed into a \textit{vector-valued} image $J: \Omega \rightarrow \mathbb{R}^N$ of its associated (local) features, with $N$ equal to the number of features. The transition from $I$ to $J$ can be formally described by means of a map $\mathcal{M}: I \mapsto J$, which is equal to identity in the case when the only features of interest are the gray-levels of $I$. In general, however, at each $\xx \in \Omega$, $J(\xx)$ may consist of, for example, the local statistics of $I(\xx)$, its associated Laws' texture features~\cite{Laws80}, multi-resolution moments~\cite{Suhling04}, or any combination thereof. The choice of features is wide and diverse; a specific combination of such features is usually selected based on the application at hand. 

Let $\Omega_m$ be a subset of $\Omega$ over which the object of interest is supported. The segmentation algorithm proposed in this paper is based on the following two assumptions.  First, it is assumed that a set of training images is available with their corresponding $\Omega_m$ being identified.  Second, it is assumed that for each $\xx \in\Omega_m$, the $N$ components of $J(\xx)$ are realizations of $N$ independent random variables. The latter assumption is  obviously an approximation, whose validity can be ameliorated by means of a whitening transform \cite{Duda01}.  

Let $P_m(\zz)$, where $\zz \in \mathbb{R}^N$, be the joint probability density function (pdf) of the image features corresponding to $\Omega_m$. Due to the assumption of statistical independence, $P_m(\zz)$ can be factorized as $P_m(\zz) = \prod_{k=1}^N p_m(z_k)$, with $p_m(z_k)$ being the pdf of the $k$-th image feature $z_k \in \mathbb{R}$. Given $\Omega_m$, the pdf $p_m(z_k)$ can be approximated according to
\begin{equation}\label{pm}
p_m(z_k) = \mathcal{E} \left\{ \frac{\int_{\Omega_m} K(z_k - J_k^{tr}(\xx)) \, d\xx}{\int_{\Omega_{m}} d\xx} \right \},
\end{equation}
where $K$ denotes a (positive valued, normalized) {\em kernel} function, $J_k^{tr}(\xx)$ is the value of the $k$-th component of a training image $J^{tr}$ at location $\xx$, and $\mathcal{E}$ stands for the operator of assemble average taken over the training set. Note that due to the normalization of $K$ (i.e. $\int_\mathbb{R} K(s) \, ds = 1$), the estimate in (\ref{pm}) is, in fact, a non-parametric kernel density estimate of the true pdf of $z_k$ \cite{Silverman86}. It should also be emphasized that the estimates $p_m(z_k)$ are assumed to be pre-computed before the actual segmentation is initialized.

Now, let $I$ be an observed image to be segmented, and $J$ be its corresponding feature image. Also, let $\Omega_{in} \subseteq \Omega$ be an arbitrary (yet non-empty) subset of $\Omega$. In this case, an approximation similar to that in (\ref{pm}) can be employed to estimate the {\em empirical} pdf $P(\zz \mid \Omega_{in})$ of the features of $I$ observed over $\Omega_{in}$. Due to the assumption of statistical independence, $P(\zz \mid \Omega_{in})$ can be factorized similarly to $P_m$, with its $N$ factors computed according to 
\begin{equation}\label{E1}
p(z_k \mid \Omega_{in}) = \frac{\int_{\Omega_{in}} K(z_k - J_k(\xx)) \, d\xx}{\int_{\Omega_{in}} d\xx}, \quad \mbox{ with } k = 1, 2, \ldots, N,
\end{equation}
so that $P(\zz \mid \Omega_{in}) = \prod_{k=1}^N p(z_k \mid \Omega_{in})$.

The ultimate goal of image segmentation is to ``deform" $\Omega_{in}$ so as to make it closely approximate the subset $\Omega_m$ that supports the object of interest. Moreover, whenever $\Omega_{in}$ coincides with $\Omega_m$, the distributions $P(\zz \mid \Omega_{in})$ and $P_m(\zz)$ should be close one to another in some sense. In this paper, as a measure of similarity between probability distributions, we use the Bhattacharyya coefficient $B$ which is defined as given by \cite{Bhattacharyya43, Kailath67, Goudail04}
\begin{equation}\label{E2}
B(\Omega_{in}) = \int_{\mathbb{R}^N} \sqrt{P_m(\zz) P(\zz \mid \Omega_{in})} \, d\zz.  
\end{equation}
Alternatively, the statistical independence of $z_k$ allows rewriting the above expression in a slightly simplified form
\begin{equation}\label{E3}
B(\Omega_{in})=\prod_{k=1}^N B_k(\Omega_{in})  = \prod_{k=1}^N \int \sqrt{p_m(z_k) \, p(z_k \mid \Omega_{in})} \, dz_k.
\end{equation}

The Bhattacharyya coefficient $B(\Omega_{in})$ achieves its maximum value of 1 when $p_m(z_k) = p(z_k \mid \Omega_{in}), \forall k$, which happens when the sets $\Omega_{in}$ and $\Omega_m$ coincide with each other. Conversely, by maximizing $B$ as a function of $\Omega_{in}$ one can reasonably expect to minimize the discrepancy between $\Omega_{in}$ and $\Omega_m$ -- the fact that forms the basis of the segmentation method of \cite{Freedman04, Freedman05}. Needless to say, the maximization of $B$ over all possible $\Omega_{in}$ is a computationally intractable problem, which happens to have an elegant solution when $\Omega_{in}$ is replaced by its implicit definition in terms of a level-set function as explained next.

\subsection{Level-set formulation}
In recent years, the level-set framework of \cite{Sethian99} has gained wide popularity in the area of image segmentation for a number of its remarkable advantages. In particular, level-set methods allow one to perform numerical computations involving curves and surfaces without the need to explicitly parameterize these objects. Thus, in the current setting, the subset $\Omega_{in}$ can be defined as  
\begin{equation}\label{E5}
\Omega_{in} = \left\{ \xx \in \Omega \mid \ls \leq 0 \right\},
\end{equation}
where $\varphi : \Omega \rightarrow \mathbb{R}$ is a {\em level-set function}, whose zero level-set $\{ \xx \in \Omega \mid \ls = 0\}$ is used to implicitly define the related active contour. 

Expressing $\Omega_{in}$ in terms of its associated $\varphi$ leads to a different definition of the Bhattacharyya coefficient, which now becomes a function of $\varphi$, {\it viz.}
\begin{equation}\label{E6}
B(\varphi) = \prod_{k=1}^N B_k(\varphi) = \prod_{k=1}^N \int \sqrt{p_m(z_k) \, p(z_k \mid \varphi)} \, dz_k,
\end{equation}
where the densities $p(z_k \mid \varphi)$ can be computed according to
\begin{equation}\label{E7}
p(z_k \mid \varphi) = \frac{\int_{\Omega} K(z_k - J_k(\xx)) \, \mathcal{H}(-\varphi(\xx))\, d\xx}{\int_{\Omega}  \mathcal{H}(-\varphi(\xx)) \, d\xx},
\end{equation}
with $\mathcal{H}(x) = (x)_+$ standing for the Heaviside function. Subsequently, the problem of finding an optimal $\Omega_{in}$ can be supplanted by an equivalent problem of finding an optimal level-set function $\varphi^\ast$ as given by
\begin{equation}\label{optim}
\varphi^\ast = {\rm arg} \max_{\varphi \in \Phi} B(\varphi),
\end{equation}
where $\Phi$ denotes a set of functions to which $\varphi$ can be formally ascribed. Thus, for example, it is common to define $\Phi$ to be a set of {\em signed distance functions} -- the choice which leads to particularly efficient numerical implementation of (\ref{optim}) \cite{Sethian99}.

Due to the absence of a closed form solution to the above problem, a numerical scheme for maximizing (\ref{E6}) is needed. In the case at hand, a standard approach to the solution of (\ref{optim}) is by means of a steepest ascent procedure which prescribes approximating $\varphi^\ast$ as a stationary point of the sequence of solutions produced by the gradient flow
\begin{equation}
\frac{\partial \varphi(\cdot, t)}{\partial t} = \frac {\delta B(\varphi)} {\delta \varphi},
\end{equation}
where a virtual (iteration) ``time" $t$ is introduced to allow the level-set function to evolve into a family $\varphi(\xx, t)$, and ${\delta B(\varphi)} \slash {\delta \varphi}$ stands for the first variation of $B$ w.r.t. the level-set function. Particularly, straightforward computation leads to
\begin{equation}\label{E8}
\frac{\delta B(\varphi)}{\delta \varphi} =  V_B \, \|\nabla\varphi \|,
\end{equation}
with $\| \nabla \varphi \|$ denoting the (Euclidean) norm of the gradient $\nabla \varphi$ of $\varphi$,  and
\begin{equation}\label{E11}
V_B(\xx) = \frac{1}{2A} \sum_{k=1}^N \alpha_k  \left( B_k(\varphi) - \Big[ r(z_k \mid \varphi) \ast K(z_k) \Big]_{z_k = J_k(\xx)} \right),
\end{equation}
where $A := \int_\Omega \mathcal{H}(-\varphi(\xx)) d\xx$, $r(z_k \mid \varphi) := \sqrt{p_m(z_k) /  p(z_k \mid \varphi)}$, $\alpha_k := \prod_{i=1, i \neq k}^N B_i(\varphi)$, and the asterisk denotes the operation of linear convolution, which can be performed using any fast algorithm.

It should be noted that the estimation of $\varphi^\ast$ has been so far performed based on regional features of the observed image $I$. This approach is particularly useful in the cases when the objects of interest do not possess well-identifiable edges \cite{ChanVese01}. However, in situations when the edges can be detected, it would be an omission not to consider such an important source of information. Technically, edge-related information can be incorporated via using the framework of {\em geodesic active contours} \cite{Caselles97, Yezzi97}. In this case, the optimal level-set function $\varphi^\star$ is estimated according to
\begin{equation}\label{E13}
\varphi^\star(\xx) = \arg \max_{\varphi \in \Phi} \left\{ \alpha \, B(\varphi) - \int_{\Omega} g(\xx)\,  \| \nabla \mathcal{H}(\varphi(\xx))\| \, d\xx \right\},
\end{equation}
where $\alpha > 0$ is a regularizing constant\footnote{In the current paper, the value of $\alpha$ has been set to be equal to 2.}, and $g : \Omega \rightarrow \mathbb{R}^+$ is an edge detector function, which can be defined as, e.g., $g = (1+ \| \nabla \tilde{I} \|^2)^{-1}$ with $\tilde{I}$ being a smoothed version of $I$. It is straightforward to show that the gradient flow corresponding to (\ref{E13}) is given by
\begin{equation}\label{E15}
\frac{\partial \varphi(\cdot, t)}{\partial t} = \left(\alpha \, V_B + {\rm div} \left( g \, \frac{\nabla \varphi}{\| \nabla \varphi \|} \right)\right) \|\nabla\varphi \|.
\end{equation}
It is worth noting that the $V_B$ term in (\ref{E15}) attempts to find a region $\Omega_{in}$ in the image domain whose empirical density $P(\zz \mid \Omega_{in})$ closely matches the model density $P_m(\zz)$ in terms of the Bhattacharyya metric. The second term, on the other hand, attempts to match the boundary of $\Omega_{in}$ to the strong edges of $I$, while maintaining minimal curvature of the related active contour.

\subsection{Tracking of shape features}
The optimization (\ref{E13}) is particularly useful in the cases when the edges of the object(s) of interest can be well-defined. Unfortunately, due to the possibility of background clutter and the presence of strong noise, using the edge information alone may not result in a sufficient regularization force, in which case the gradient flow of (\ref{E15}) needs to be subjected to additional constraints. A common solution to the aforementioned problem lies in the concept of shape priors \cite{Leventon00, Chen02, Tsai03, Chan03, Freedman05}. In this paper, as a means to distinguish the proposed ``weak" shape priors and the above mentioned methods, the latter are considered as based on ``strong" shape priors, which impose a substantial restriction on the evolution of active contours. As will be demonstrated shortly, while highly desirable in certain scenarios, such ``strong" priors can bias the segmentation towards a suboptimal (or even useless) solution in situations when the training set consists of a relatively small number of examples. The ``weak" shape priors, on the other hand, are less susceptible to the above limitation which, in combination with their property of being invariant under a group of geometric transformations, makes the proposed priors a viable alternative to the existing shape models.  

As a means to create a ``weak" shape model based on the distribution tracking framework, one must first extract a morphological feature (or a set thereof) which will be used to characterize the boundary of the object of interest. To this end, let $\varphi^m$ be the level-set function corresponding to a known $\Omega_m$, and let $\Gamma_m = \{ \xx \in \Omega \mid \varphi^m(\xx) = 0 \}$ be the boundary of $\Omega_m$. In the current paper, the ``weak" shape priors are constructed using the {\em curvature} of $\Gamma_m$ as a morphological feature. Note that the curvature has been chosen primarily for its property of being invariant under the group of Euclidean transformations -- the property that allows the algorithm to forgo the preprocessing step of alignment and registration via rotations and translations. It should be noted, however, that the proposed approach is by no means restricted to the tracking of curvature alone, as the very same methodology could be applied to other geometric descriptors of curves, should one require invariance under a different class of  geometric transformations \cite{Sapiro93, Bruckstein97}.

In the case when $\varphi^m$ is defined to be a signed distance function \cite{Sethian99}, the values of the curvature of $\Gamma_m$ can be obtained from the values of the curvature $\kappa_m$ of the level sets of $\varphi^m$ computed according to 
\begin{equation}\label{E16}
\kappa_m = - {\rm div} \left\{ \frac{\nabla \varphi^m}{\| \nabla \varphi^m \|} \right\}.
\end{equation}
Thus, given a training set of segmented images, the {\em model} pdf of the object curvature can be estimated as given by
\begin{equation}\label{cpdf}
C_m (\eta) = \mathcal{E} \left\{ \frac{ \int_\Omega \delta_\epsilon(\varphi^m(\xx)) \, K(\eta - \kappa_m(\xx)) \, d\xx}{ \int_\Omega \delta_\epsilon(\varphi^m(\xx)) \, d\xx} \right\},
\end{equation}
where $\mathcal{E}$ denotes the operator of assemble average as before, and $\delta_\epsilon$ denotes a smoothed delta function, which can be defined, e.g., as \cite{ChanVese01}
\begin{equation}\label{E10}
\delta_\epsilon(x) =
\begin{cases}
(2 \epsilon)^{-1} \left(1 + \cos(\pi x \slash \epsilon)\right), &|x| \leq \epsilon \\
0, &{\rm otherwise}.
\end{cases}
\end{equation}
Note that the role of $\delta_\epsilon$ in (\ref{cpdf}) is to ``weight" the domain $\Omega$ so as to make the estimated pdf depend on the values of $\kappa_m$ in close proximity of the active contour. (In our experimental study, the parameter $\epsilon$ is equal to 2.) It is also worthwhile noting that the estimate in (\ref{cpdf}) is conceptually analogous to {\em weighted} kernel density estimation \cite{Duda01}. Just as in the case of the ``photometric" densities, the model pdf $C_m(\eta)$ is supposed to be pre-computed and stored before the actual segmentation is initialized.

Given a level set function $\varphi$, the {\em empirical} probability density $C(\eta \mid \varphi)$ of the curvature of its associated active contour can be estimated according to
\begin{equation}\label{E17}
C(\eta \mid \varphi) = \frac{\int_\Omega \delta_\epsilon(\ls) \, K\left(\eta + {\rm div\{ \nabla \ls \slash \| \nabla \ls\| \}}  \right) \, d\xx}{\int_\Omega \delta_\epsilon(\ls) \, d\xx}.
\end{equation}
Subsequently, the similarity between the model and empirical densities can again be assessed using the Bhattacharyya coefficient, which in this case is given by
\begin{equation}\label{E18}
B_c(\varphi) = \int \sqrt{C_m(\eta) \, C(\eta \mid \varphi)} \, d\eta,
\end{equation} 
where subscript $c$ has been added as a means to discriminate between the above coefficient and the one in (\ref{E6}).

Analogously to the tracking of photometric variables, we propose to track the empirical density $C(\eta \mid \varphi)$ by forcing it to be as similar as possible to the model density $C_m(\eta)$ through maximization of $B_c(\varphi)$ in (\ref{E18}). The maximization can be achieved using the same tool of steepest ascent in the direction of the first variation of $B_c(\varphi)$ computed w.r.t. $\varphi$. Under the assumption on $\varphi$ to be a signed distance function (which implies $\|\nabla \ls\| = 1, \forall \xx$), the above first variation is given by
\begin{align}\label{E19}
V_c(\xx) := \frac{\delta B_c(\varphi)}{\delta\varphi} = &\frac{1}{2} \Bigg[ \int L(\eta) \, \Delta\left[\delta_\epsilon(\varphi) \, K^\prime(\eta-\kappa(\xx))\right]d\eta + \notag\\
&+ \delta_\epsilon^\prime(\varphi) \Big( \big[ L(\eta) \ast K(\eta) \big]_{\eta = \kappa(\xx)} - B_c(\varphi)\Big) \Bigg],
\end{align}
where the prime stands for differentiation w.r.t. $\eta$, $\kappa(\xx) = - {\rm div\{ \nabla \ls \slash \| \nabla \ls\| \}}$, $L(\eta) = \sqrt{ C_m(\eta) \slash C(\eta \mid \varphi) }$, and $\Delta$ is the operator of Laplacian. Subsequently, the optimization problem of (\ref{E13}) can be extended to incorporate the additional shape information, which results in a new optimal $\varphi^\ast$ computed according to
\begin{equation}\label{E20}
\varphi^\ast = \arg \max_{\varphi \in \Phi} \left\{ \alpha \, B(\varphi) + \beta \, B_c(\varphi) - \int_{\Omega} g(\xx)\,  \| \nabla \mathcal{H}(\ls)\| \, d\xx \right\},
\end{equation}
where $\Phi$ denotes the set of signed distance functions (which can be identified with the set of solutions of the eikonal equation $\| \nabla \varphi\| = 1$, subject to $\ls \big|_{\xx \in \partial \Omega_{in}} = 0 $), and $\beta > 0$ is another regularization parameter, which is set to be equal to 5 in the experimental study of this paper. Finally, the gradient flow that corresponds to the maximization problem (\ref{E20}) is given by
\begin{equation}\label{E21}
\frac{\partial \varphi(\cdot, t)}{\partial t} = \| \nabla \varphi \| \left( \alpha \, V_B + {\rm div} \left( g \frac{\nabla \varphi}{\| \nabla \varphi \|} \right) \right) + \beta \, V_c.
\end{equation}
In the present study, the gradient flow in (\ref{E21}) has been implemented using an implicit discretization scheme based on the method of additive operator splitting (AOS) as detailed in \cite[Ch. 3 - 4]{Osher03}.

\section{Numerical Considerations}

\begin{figure}[top]
\centering

\makebox[\textwidth]{\includegraphics[width=6in]{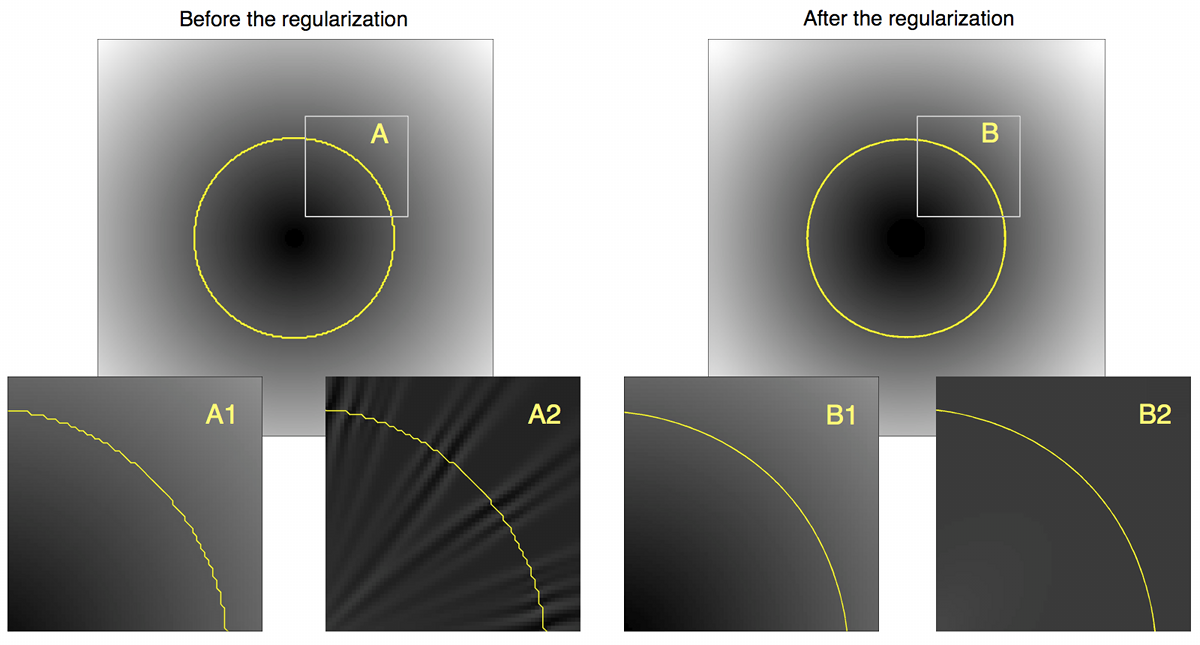} }
\caption{(Subplot A) A level-set function and its corresponding zero level-set before regularization. (Subplot A1) A zoomed section of the level-set function in Subplot A. (Subplot A2) The curvature of the level-sets in Subplot A1. (Subplot B) The regularized version of the level-set function in Subplot A. (Subplot B1) A zoomed portion of the regularized level-set function in Subplot B. (Subplot B2) The curvature of the level-sets in Subplot B1.}  
\label{F1}
\end{figure}

\subsection{Regularization of curvature}
In numerical computations, the values of the curvature $\kappa$ of the level-sets of $\varphi$ are commonly computed according to
\begin{equation}\label{E22}
\kappa = - {\rm div} \left\{ \frac{\nabla \varphi }{\| \nabla \varphi \|} \right\} = \frac{2\varphi_x\varphi_y\varphi_{xy}-\varphi_{xx}\varphi_y^2-\varphi_{yy}\varphi_x^2}{(\varphi_x^2+\varphi_y^2)^{3/2}},
\end{equation}
where subscripts $x$ and $y$ denote partial differentiation along the respective directions. Since the partial derivatives in (\ref{E22}) are standardly approximated by means of finite-support numerical schemes, the practical values of $\kappa$ should be expected to be flawed by quantization noises \cite{Macklin06}. Consequently, the empirical pdf of $\kappa$ will be a poor estimate of the original curvature, if the effect of the noises is not appropriately dealt with. The problematic aspect of the above argument is exemplified in Fig.~\ref{F1}, where Subplot A shows a signed distance function (visualized as a grayscale image) corresponding to a circle that is depicted in yellow in the same subplot. In addition, Subplot A1 shows a zoomed section of Subplot A (where the effect of quantization can be clearly observed), while Subplot A2 superimposes the circle over an image of its related curvature $\kappa$. One can see that the approximated values of $\kappa$ suffer from spurious variations, which contradict the theoretical behavior of $\kappa$. 

In this paper, to alleviate the problem of irregularity of the discrete approximation of $\kappa$, we propose to subject the level-set function $\varphi(\cdot, t)$ to a pre-smoothing procedure, before $\varphi(\cdot, t)$ is used to estimate the pdf of $\kappa$ according to (\ref{E17}). Specifically, the pre-smoothing can be achieved via {\em anisotropically} diffusing $\varphi$ by means of the following equation \cite{Weickert99} 
\begin{equation}\label{E23}
\frac{\partial\varphi(\xx, \tau)}{\partial \tau} = {\rm div} (D(\xx)\,\nabla\varphi(\xx, \tau)),
\end{equation}
where $D(\xx) \in \mathbb{R}^{2 \times 2}$ denotes a diffusivity tensor at $\xx$, and $\tau$ denotes an artificial diffusion time (which is not to be confused with the one in (\ref{E21})).

The smoothing effect of the diffusion (\ref{E23}) is controlled through the spectral properties of $D(\xx)$ which, in the case at hand, should be defined in such a way that the smoothing will propagate along the directions {\em tangent} to the level-sets of $\varphi$. Note that diffusing in the tangential direction allows preserving the shape of the active contour, while effectively suppressing the spurious irregularities in $\varphi$ which have been induced by discretization. In particular, the above objective can be achieved via setting $D(\xx) = \gamma \, \vec{v}_1(\xx) \vec{v}_1(\xx)^T + \vec{v}_2(\xx) \vec{v}_2(\xx)^T$, with $\gamma \ll 1$, and with $\vec{v}_1(\xx)$ and $\vec{v}_2(\xx)$ being two unit vectors pointing in the directions parallel and perpendicular to $\nabla \ls$, respectively. Namely, $\vec{v}_1(\xx) \parallel \nabla\ls$ and $\vec{v}_2(\xx) \perp \nabla \ls$ at every $\xx \in \Omega$.   

In the experiments reported in this paper, the diffusion in (\ref{E23}) was performed using the AOS scheme as detailed in \cite{Weickert99}, with $\gamma = 0.01$, and with the number of iterations and the diffusion time-step being equal to $4$ and $\Delta \tau = 5$, respectively. The advantage of the proposed  preprocessing stage can be appreciated via observing Subplots B1 and B2 of Fig.~\ref{F1}, which show a regularized version of the level-set function of Subplot A and its associated curvature, correspondingly. One can see that the preprocessing brings the discrete values of $\kappa$ to a closer correspondence with its theoretically predicted values. 

\subsection{Numerical algorithm}

For the sake of reproducibility, a pseudocode of the proposed algorithm is summarized in Algorithm 1 below. Input into the algorithm consists of the original image $I$, a model distribution of its photometric features $P_m$, a model curvature distribution $C_m$, as well as of an initial level-set function $\varphi_0$. The output of the algorithm consists of an optimal level-set function $\varphi^\ast$, whose zero level-set defines the boundary of the object of interest.

\begin{algorithm}[H]
\caption{Proposed segmentation procedure} 
\begin{algorithmic}[1]
\STATE {\textbf{given:}} $I$, $P_m$, $C_m$, $\varphi_0(\cdot) \equiv \varphi(\cdot,t=0)$
\STATE {\textbf{preset:}} $\Delta t = 5$, $\alpha =2$, $\beta = 5$, $t=0$
\STATE {\textbf{compute:}} $\{J_k\}_{k=1}^N$ and an edge detector function $g$
\WHILE {$\delta > 10^{-3}$}
\STATE {Diffuse $\varphi^{(t)}$ using (\ref{E23}) to result in $\tilde{\varphi}^{(t)}$}
\STATE {Compute $\kappa = - {\rm div} \left\{ \nabla \tilde{\varphi}^{(t)} \slash \| \nabla \tilde{\varphi}^{(t)} \| \right\}$ using (\ref{E22})}
\STATE {Compute $\left\{ p(z_k \mid \varphi^{(t)}) \right\}_{k=1}^N$ using (\ref{E7}) and $C(\eta \mid \varphi^{(t)})$ using (\ref{E17})} 
\STATE {Compute $V_B$ using (\ref{E11}) and $V_c$ using (\ref{E19})}
\STATE {$\varphi^{(t+1)} \Leftarrow \varphi^{(t)} + \Delta t \, \left( \alpha \, V_B + \beta \, V_c \right)$}
\STATE {$\varphi^{(t+1)} \Leftarrow {\rm AOS}\left(\varphi^{(t+1)}, g,\Delta t\right)$}
\STATE {Redistance $\varphi^{(t+1)}$ by fast marching \cite{Sethian99}}
\STATE {$\delta \Leftarrow \| \varphi^{(t+1)} - \varphi^{(t)} \|$}
\STATE {$t \Leftarrow t+1$}
\ENDWHILE
\RETURN {$\varphi^\ast = \varphi^{(t)}$}
\end{algorithmic}
\end{algorithm} 

\section{Experimental Results}

\subsection{Segmentation evaluation metrics}\label{metr}

\begin{figure}[top]
\centering
\makebox[\textwidth]{ \includegraphics[width=5.5in]{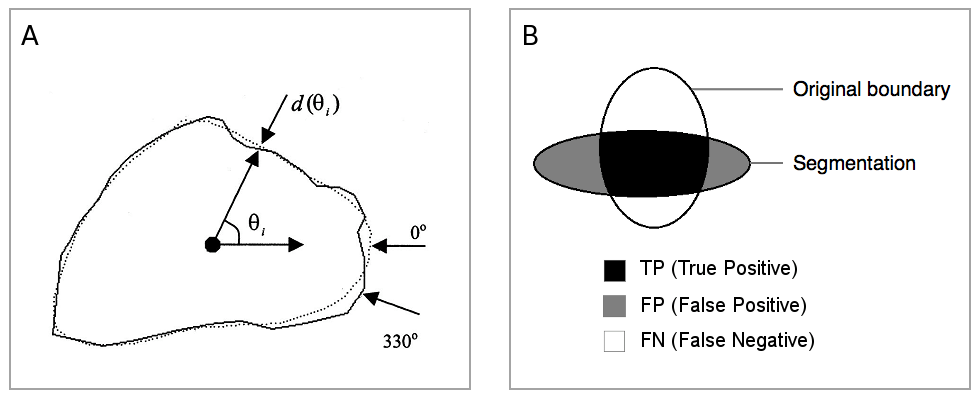} }
\caption{ (Subplot A) Local difference $d(\theta_i)$ between the original and optimal boundaries; (Subplot B) Different regions within the original and optimal segmentations.}
\label{F3}
\end{figure}

In this section, to qualitatively compare the performance of the proposed and reference segmentation methods, a number of comparison metrics are employed. Specifically, given the true boundary of a desired object and its estimate, both contours can be sampled at the points defined by their intersection with $L$ rays emanating from a common barycenter of the contours (see Subplot A of Fig.~\ref{F3} for an example). Subsequently, along the directions of the rays (defined by angles $\theta_i = 2\pi i/L$, with $i=0,1, \ldots, L$), the signed distances $\{d(\theta_i)\}_{i=1}^L$ between the original and optimal points can be computed and used to compute mean difference (MD), mean absolute difference (MAD), and maximum difference (MAXD) defined as \cite{Ladak00}
\begin{subequations}
\begin{equation}
{\rm MD} = \frac{1}{L} \sum_{i=1}^L d(\theta_i),
\end{equation}
\begin{equation}
{\rm MAD} = \frac{1}{L} \sum_{i=1}^L | d(\theta_i) |,
\end{equation}
\begin{equation}
{\rm MAXD} = \max_{1 \le i \le L} \{ | d(\theta_i) | \}.
\end{equation}
\end{subequations}
Whereas $\rm MD$ provides useful information on the size and direction of the segmentation bias, $\rm MAD$ and $\rm MAXD$ quantify the overall segmentation error in terms of the average and maximum value of $\{| d(\theta_i) |\}_{i=1}^L$, respectively.

Additional performance measures can be derived based on area-based metrics, which (as opposed to MD, MAD, and MAXD) are insensitive to the local geometry of segmentation boundaries. Such metrics can be defined using the nomenclature of detection theory, in which case they are computed based on the differences between the original and optimal segmentations viewed as subsets of the image domain $\Omega$. Particularly, let {\em true positive}, {\em false positive}, and {\em false negative} subsets of $\Omega$ be defined as shown in Subplot B of Fig.~\ref{F3}, with their corresponding areas denoted by TP, FP, and FN, respectively. Then, following \cite{Ladak00}, one can define three area-based metrics as given by
\begin{subequations}\label{meas}
\begin{equation}
{\rm SEN} = \frac{{\rm TP}} {{\rm TP} + {\rm FN}},
\end{equation}
\begin{equation}
{\rm ACC} = 1 - \frac{{\rm FP} + {\rm FN}} {{\rm TP} + {\rm FN}},
\end{equation}
\begin{equation}
{\rm AO} = \frac{{\rm TP}}{{\rm TP} + {\rm FP} + {\rm FN}}.
\end{equation}
\end{subequations}
With TP + FN being equal to the area of the original segmentation, the SEN, ACC, and AO measures quantify the sensitivity, accuracy and area-overlap of the optimal segmentation, respectively \cite{Ladak00}. It is worthwhile noting that, in the case of errorless segmentation, all the measures in (\ref{meas}) attain their maximum value of 1. Thus, in general, the higher the values of SEN, ACC, and AO, the better the performance of an image segmentation algorithm under consideration. 

Finally, the segmentation accuracy can be also assessed in terms of the the standard {\it mean squared error} (MSE) criterion. Specifically, let $F$ be the $M \times N$ array of the true (binary) segmentation mask, and $\tilde{F}$ be its estimate. Then, the MSE is defined as
\begin{equation}\label{E24}
{\rm MSE} = \frac{1}{MN} \sum_{x=0}^{M-1} \sum_{y=0}^{N-1} \left| \tilde{F}(x,y)-F(x,y) \right|^2.
\end{equation}

\subsection{Experiments with MPEG shape data}\label{MPEG}
In this section, shape data from the MPEG-7 core experiments CE-Shape-1 database \cite{Latecki00} is exploited to validate the performance of the proposed segmentation method. In particular, the ``face" and ``teddy" data-sets are used throughout this section. From the set of 20 images in each data-set, a total of 20 leave-one-out validation experiments were performed for which 19 images were used for training, while the remaining image was used for testing. Hence, the comparative figures summarized in Table~\ref{T1} below have been obtained via averaging the results of 20 independent trials.

\begin{figure}[top]
\centering
\makebox[\textwidth]{\includegraphics[width=5in]{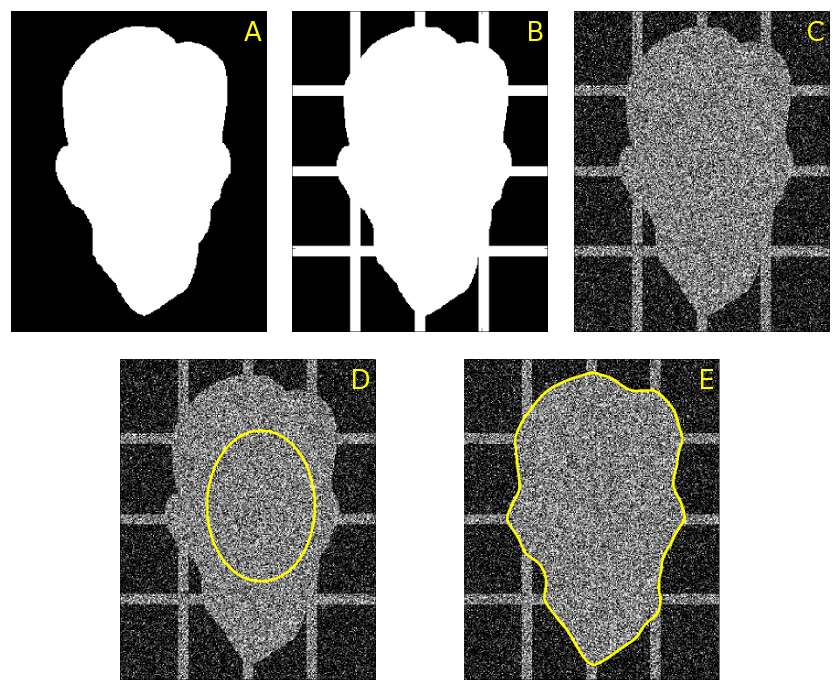}}
\caption{Segmentation of a ``face" image with compromised edges. (Subplot A) Original binary image;(Subplot B) Original binary image with added clutter; (Subplot C) Cluttered image contaminated with Gaussian noise; (Subplot D) Initial segmentation; (Subplot E) Final segmentation.}
\label{F5}
\end{figure}

\begin{figure}[top]
\centering
\makebox[\textwidth]{ \includegraphics[width=5in]{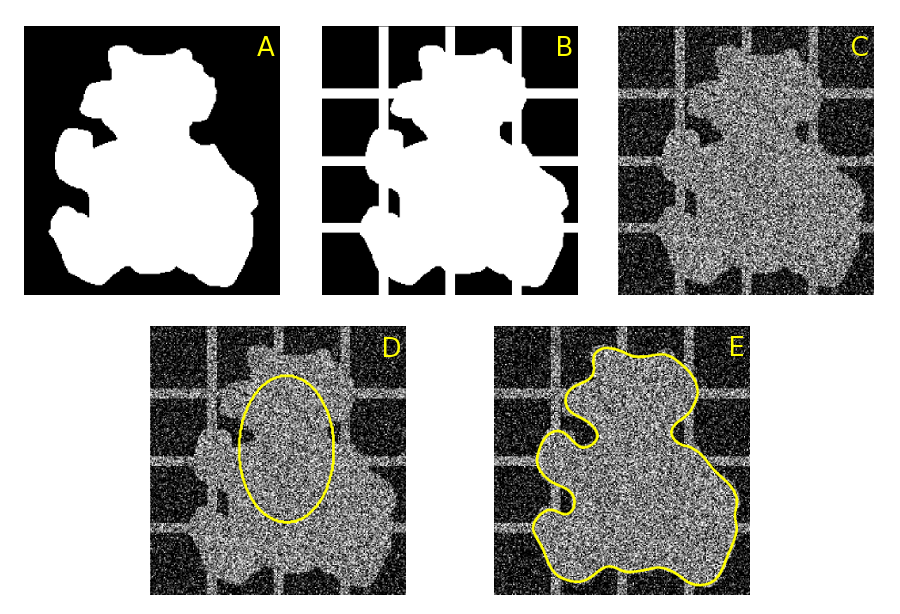} }
\caption{Segmentation of a ``teddy" image with compromised edges. (Subplot A) Original binary image;(Subplot B) Original binary image with added clutter; (Subplot C) Cluttered image contaminated with Gaussian noise; (Subplot D) Initial segmentation; (Subplot E) Final segmentation.}
\label{F6}
\end{figure}

Each segmentation experiment was preceded by a training stage, during which a total of 19 original images in each data-set were used to learn the model densities $P_m(\zz)$ and $C_m(\eta)$, while the remaining $20^{\rm th}$ image (used for the actual segmentation) was intentionally left out\footnote{Note that, in this case, the expectations in (\ref{pm}) and (\ref{cpdf}) are approximated by the assemble average of 19 probability densities.}. The photometric pdf $P_m(\zz)$ was computed based on two image features, {\em viz.} the values of a data image $I$ as well as of its linearly smoothed version $\tilde{I}$, i.e. $J_1 = I$ and $J_2 = \tilde{I}$. The data images, in turn, were obtained from the original images by contaminating them with both horizontal and vertical line clutter, supplemented by the addition of white Gaussian noise to further compromise the edges (see Subplots C in Fig.~\ref{F5} and Fig.~\ref{F6} for some typical examples of resulting images).

As shown in Fig.~\ref{F5} and Fig.~\ref{F6}, the horizontal and vertical line clutters along with Gaussian noise contamination result in missing edges in both the ``face" and ``teddy" images. Needless to say, these missing edges would cause problems for segmentation algorithms which do not rely on prior shape information. The proposed algorithm, on the other hand, does not have any difficulties converging to a close approximation of the true shape. The observed regularization occurs due to the $V_c$ force as defined by (\ref{E19}); if the active contour were to ``leak" out at the cluttered areas, the empirical pdf $C(\eta \mid \varphi)$ would increasingly deviate from the model distribution $C_m(\eta)$. Thus, even though the shape prior ``encoded" in terms of a pdf distribution is considered a ``weak" prior, it appears to be sufficiently restrictive to effectively regularize the segmentation. 

\begin{table}[htbp]
\caption{Segmentation results pertaining to Section~\ref{MPEG}}
\centering
\noindent\makebox[\textwidth]{
\begin{tabular}{ l | l l | l l}
\toprule[1.5pt]
	 	&\multicolumn{2}{c |}{Face} 	&\multicolumn{2}{c}{Teddy} \\
Method	&Weak Priors ($\pm \sigma$) 	&Strong Priors ($\pm \sigma$) 
		&Weak Priors ($\pm \sigma$) 	&Strong Priors ($\pm \sigma$)\\
\midrule
MD	    	& ${\bf 0.535}\pm 0.189$ 	& $0.548\pm 0.823$ 	& ${\bf 0.758}\pm 0.568$ 	& $-1.45\pm 1.06$\\
MAD   	& ${\bf 2.240}\pm 0.101$ 	& $2.410\pm 1.230$ 	& ${\bf 2.970}\pm 0.383$ 	& $4.02\pm 1.18$\\
MAXD 	& $4.41\pm 0.39$		& ${\bf 3.91}\pm 0.44$ & ${\bf 4.66}\pm 0.35$ 	& $4.85\pm 0.25$\\
SEN	    	& ${\bf 0.974}\pm 0.001$ 	& $0.971\pm 0.016$ 	& $0.942 \pm 0.005$ 	& ${\bf 0.97}\pm 0.015$\\
ACC 	& ${\bf 0.959}\pm 0.002$	& $0.953\pm 0.022$ 	&${\bf 0.936}\pm 0.005$ 	& $0.93\pm 0.024$\\
AO 		& ${\bf 0.960}\pm 0.001$	& $0.954\pm 0.021$ 	&${\bf 0.937}\pm 0.005$ 	& $0.93\pm 0.022$\\
MSE	& ${\bf 0.017}\pm 0.001$ 	& $0.021\pm 0.009$ 	&${\bf 0.020}\pm 0.002$ & $0.023\pm 0.008$\\
\bottomrule
\end{tabular}
\label{T1}}
\end{table}

To prove that the proposed method constitutes a viable and useful alternative to existing segmentation techniques, its performance should be compared with that of a well-established reference algorithm. Selecting such an algorithm is obviously a difficult task in view of the availability of a plethora of powerful candidates. However, as the primary contribution of the present paper is in introducing a new type of shape priors, it seems reasonable to require the reference method to be identical to the proposed one, except for the shape modeling part. Consequently, in this paper, the method of \cite{Freedman05} was used for comparisons. Note that the only difference between the proposed method and the one in \cite{Freedman05} is that the shape modeling employed by the latter is based on the PCA-based approach of \cite{Leventon00}, which will be referred below to as using ``strong" shape priors.

\begin{figure}[top]
\centering
\makebox[\textwidth]{\includegraphics[width=5in]{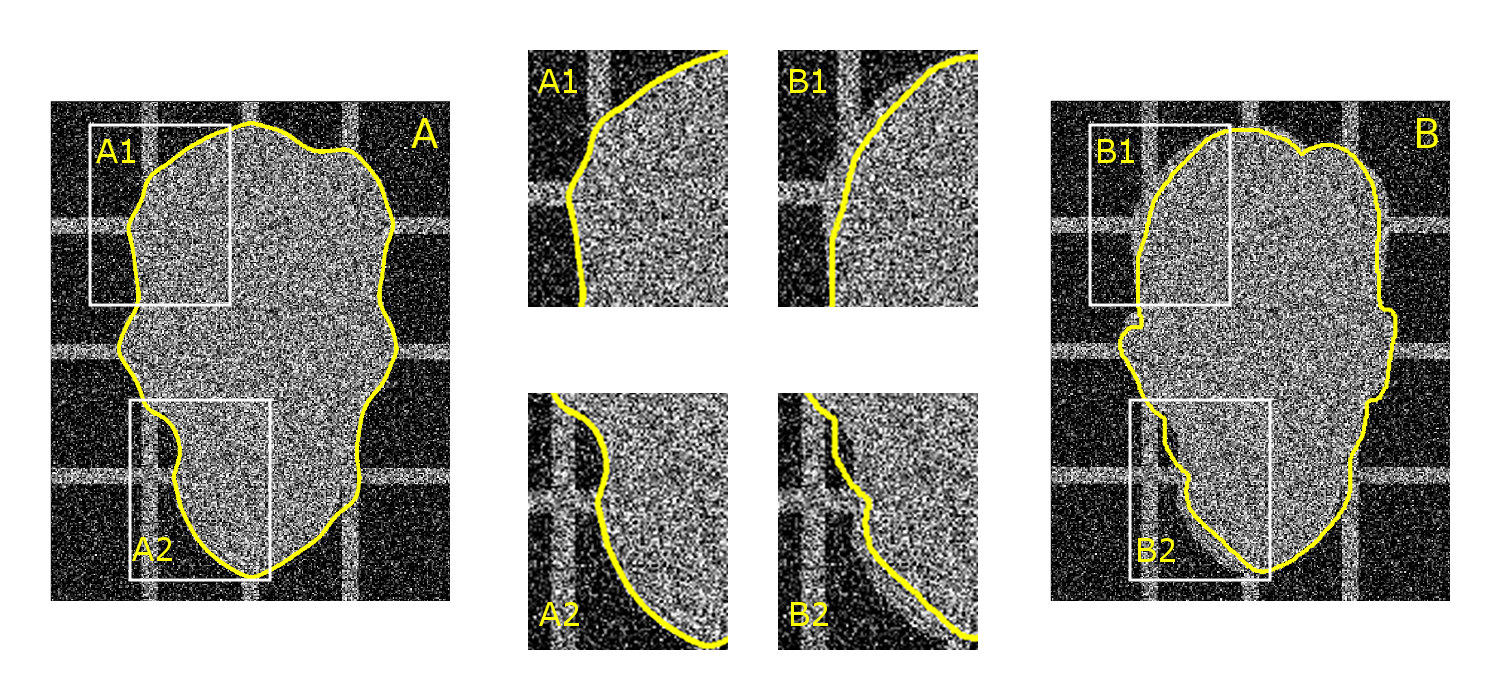} }
\caption{(Subplot A) Segmentation using ``weak" shape priors; (Subplot A1 and Subplot A2) Zoomed-in sections of the segmented ``face" image in Subplot A; (Subplot B) Segmentation using ``strong" shape priors; (Subplot B1 and Subplot B2) Zoomed-in sections of the segmented ``face" image in Subplot B.}
\label{F7}
\end{figure}

\begin{figure}[htbp]
\centering
\makebox[\textwidth]{\includegraphics[width=5in]{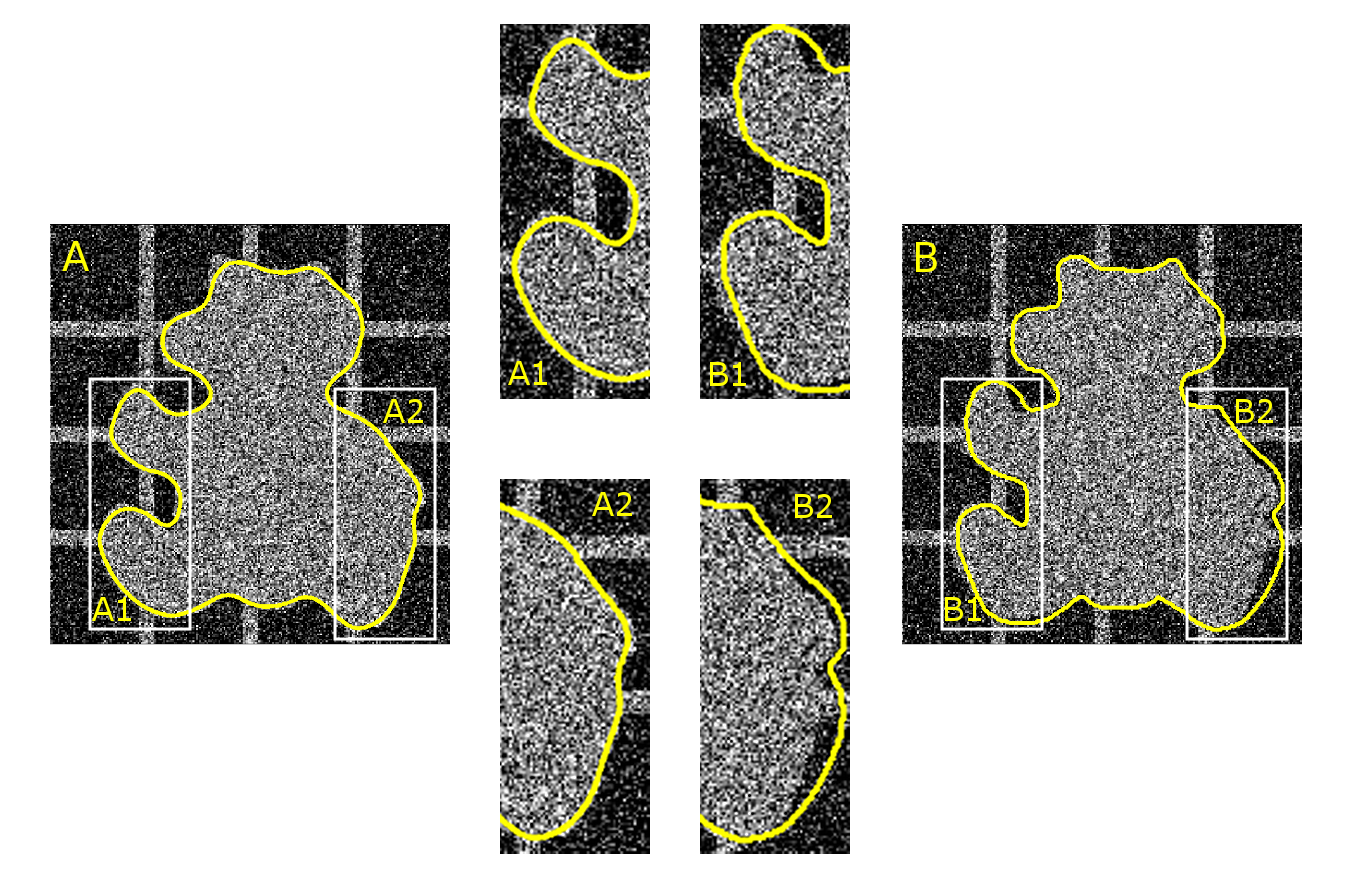} }
\caption{(Subplot A) Segmentation using ``weak" shape priors; (Subplot A1 and Subplot A2) Zoomed-in sections of the segmented ``teddy" image in Subplot A; (Subplot B) Segmentation using ``strong" shape priors; (Subplot B1 and Subplot B2) Zoomed-in sections of the segmented ``teddy" image in Subplot B.}
\label{F8}
\end{figure}

Analogously to the above case, the prior models of the reference method were learned using the ``leave-one-out" approach, with 19 images used for training and one image used for validation. A typical segmentation produced by the proposed method is depicted in Subplot A of Fig.~\ref{F7}, while Subplot B of the same figure shows the segmentation produced by the reference method. One can see that the proposed method is able to accurately delineate the ``face" image as can be evidenced in the zoomed-in portions of the image depicted in Subplot A1 and A2. On the other hand, Subplots B1 and B2 reveal estimation bias in the results obtained using the reference method. Similar segmentation results for the ``teddy" image are depicted in Fig.~\ref{F8}. Here again the segmentation obtained based on the ``weak" shape model outperforms the segmentation obtained using the ``strong" priors. The above observations and conclusions are further supported by the quantitative comparisons of Table~\ref{T1}. Particularly, in the set of ``face" test images, the ``weak" priors outperform the ``strong "priors in all the performance metrics except for the MAXD measure. Similarly, in the set of ``teddy" test images, the ``weak" priors outperform the ``strong" priors in all the performance metrics except in the SEN category.

\subsection{Sensitivity to shape outliers}\label{outliers}
The effectiveness of PCA-based prior modeling \cite{Leventon00, Tsai03, Freedman05} is known to be dependent on the size of training sets used, with larger sets resulting in more robust and reliable segmentation. However, practical situations are common when access to an abundance of training shapes may not be available. Consequently, in such cases, the performance of PCA-based shape modeling is likely to deteriorate. Furthermore, as a general rule, the smaller the size of a training set is, the poorer is the robustness of PCA-based  modeling towards erroneous training samples (outliers), which should be expected in practice. The proposed ``weak" modeling, on the other hand, are substantially less sensitive to the influence of outliers, and it requires substantially fewer training shapes to produce reliable shape models as demonstrate by our next experiment.  

In order to quantitatively compare the robustness of the ``weak" and ``strong" shape models, the same data-sets consisting of ``face" and ``teddy" images was used. In this case, however, the size of the training sets was reduced from 19 to 4 sample images. Moreover, each training set was ``spoiled" by supplementing it with an outlier image of an elliptic blob (whose area was set to be approximately equal to the area of a ``face"). Subsequently, a total of 5 training images were used to learn the ``weak" and ``strong" shape models of the test images. Note that, analogous to the case of Section~\ref{MPEG}, the data images used in actual segmentation have been corrupted by both vertical and horizontal linear clutter and white Gaussian noise. 

\begin{figure}[top]
\centering
\includegraphics[width = 5in]{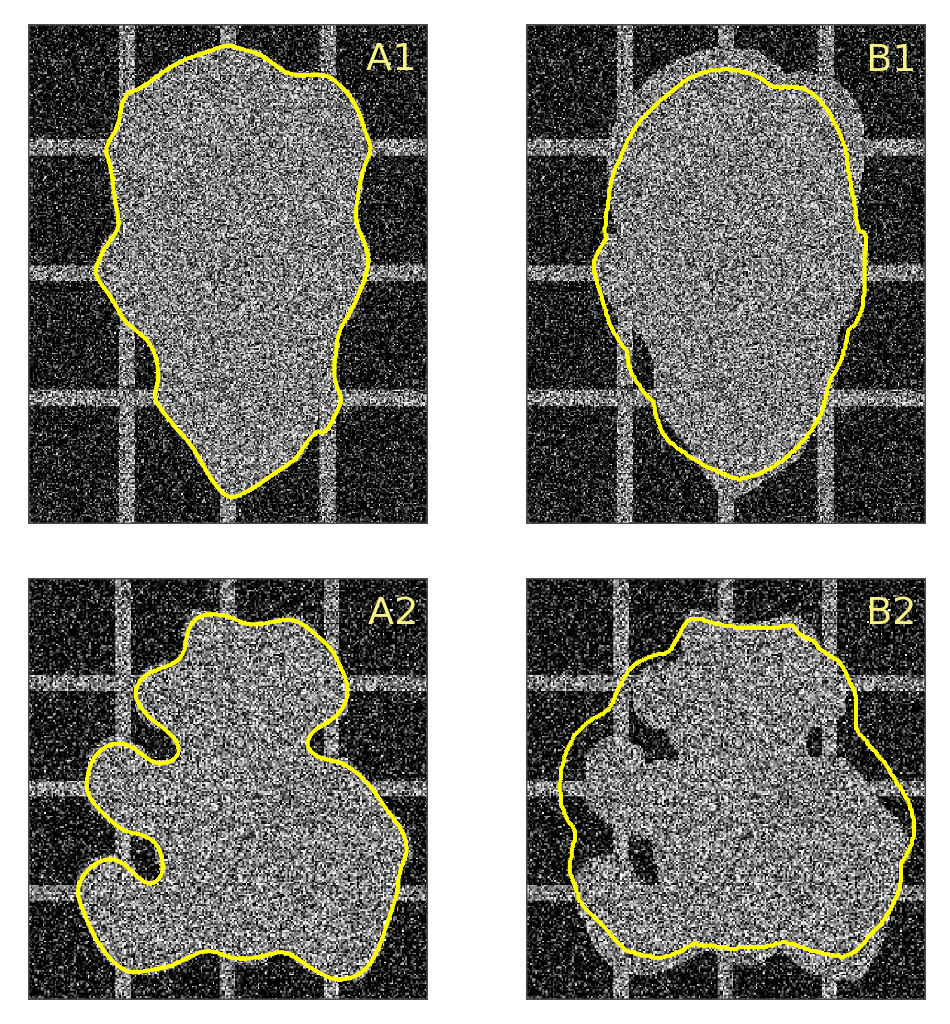} 
\caption{ (Subplot A1 and Subplot A2) Segmentations results for ``face" and ``teddy" images obtained with the proposed algorithm. (Subplot B1 and Subplot B2) Segmentation results for the same images obtained using the reference method of \cite{Freedman05}. In both cases the training data-sets consist of four correct images and one outlier.}
\label{F9}
\end{figure}

Fig.~\ref{F9} demonstrates a typical result of the current experiment. Particularly, Subplots A1 and A2 of the figure show the segmentations computed using the proposed ``weak" priors, in which case the presence of outliers does not seem to have any effect on the performance of the algorithm. The PCA-based ``strong" modeling, on the other hand, results in useless segmentation (as depicted by Subplots B1 and B2 of Fig.~\ref{F9}) which supports the above concern regarding its being susceptible to erroneous training samples. This fact is also supported by the quantitative metrics of Table~\ref{T2}, which clearly demonstrate the superiority of the proposed method over the reference one.

\begin{table}[htbp]
\caption{Segmentation results pertaining to Section~\ref{outliers}}
\centering
\noindent\makebox[\textwidth]{
\begin{tabular}{ l | l l | l l}
\toprule[1.5pt]
	 	&\multicolumn{2}{c |}{Face} 	&\multicolumn{2}{c}{Teddy} \\
Method	&Weak Priors ($\pm \sigma$) 	&Strong Priors ($\pm \sigma$) 
		&Weak Priors ($\pm \sigma$) 	&Strong Priors ($\pm \sigma$)\\
\midrule
MD 		& ${\bf 0.571}\pm 0.178$	& $1.46\pm 0.872$	& ${\bf 0.784}\pm 0.556$ 	& $-12.4\pm 1.13$\\
MAD 	& ${\bf 2.25}\pm 0.110$	& $6.15\pm 0.965$	& ${\bf 2.99}\pm 0.378$	& $18.5\pm 1.01$\\
MAXD	& ${\bf 4.53}\pm 0.39$	& $7.45\pm 0.72$   	& ${\bf 4.66}\pm 0.342$	& $5.42\pm 0.68$\\
SEN		& ${\bf 0.974}\pm 0.001$	& $0.916\pm 0.017$	& ${\bf 0.941}\pm 0.005$	& $0.91\pm 0.012$\\
ACC 	& ${\bf 0.958}\pm 0.002$ 	& $0.884\pm 0.018$	& ${\bf 0.935}\pm 0.005$ & $0.725\pm 0.019$\\
AO 		& ${\bf 0.960}\pm 0.003$	& $0.888\pm 0.017$	& ${\bf 0.936}\pm 0.004$ & $0.768\pm 0.014$\\
MSE	& ${\bf 0.017}\pm 0.001$ 	& $0.072\pm 0.099$	& ${\bf 0.021}\pm 0.002$& $0.087\pm 0.006$\\
\bottomrule
\end{tabular}
\label{T2}}
\end{table}

\subsection{Partial occlusions}\label{occl}
\begin{figure}[top]
\centering
\includegraphics[width=4.5in]{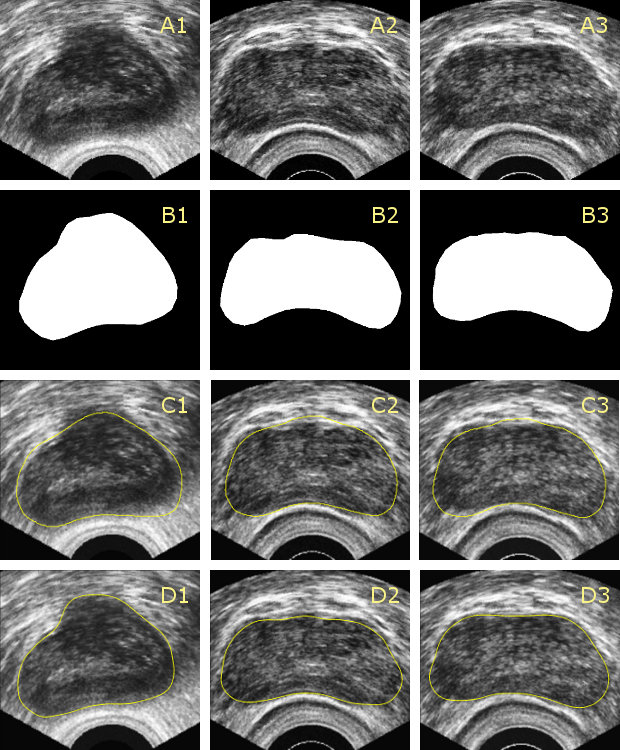}
\caption{ (Subplots A1-A3) Original TRUS images of the prostate; (Subplots B1-B3) Manual segmentations of the prostates; (Subplots C1-C3) Segmentation results produced based on the ``strong" shape model; (Subplots D1-D3) Segmentation results produced using the proposed approach with $\beta =2$.}
\label{F10}
\end{figure}

The experiments described in the preceding sections of the paper have been based on synthetic data images. In this section, the performance of both the proposed and reference algorithms is assessed using the real-life, transrectal ultrasound (TRUS) images. In general, ultrasound imaging is known to be an integrate part of modern image-based diagnostics, where it is distinguished from alternative imaging modalities because of its high benefit-to-cost ratio. Unfortunately, numerous advantages of ultrasound imaging (e.g. non-invasivenss, portability, cost efficiency, etc.) are counterbalanced by its relatively poor resolution and contrast, as compared with MRI and X-ray CT. Moreover, ultrasound images are also known to suffer from refraction and shadowing artifacts, which effectively {\em occlude} the boundaries of studied organs, thereby substantially complicating the process of their automatic segmentation. These artifacts can be observed in Subplots A1-A2 of Fig.~\ref{F10} which show three examples of TRUS images, while Subplots B1-B3 of the figure show the shapes of the corresponding prostate glands which have been manually delineated by a radiologist at the University Hospital of London (Ontario, Canada). In what follows, the results of manual segmentation will be used as a reference, against which the performance of segmentation methods will be compared.

In the experiments of this subsection, a set consisting of six TRUS images was used for training and validation. The photometric features used by the proposed and reference algorithms were defined to be the gray-level values of the TRUS images together with the values of their de-speckled versions computed by the SRAD filter of \cite{Yu02}. As in the previous experiments, curvature has been employed once again as a geometric shape descriptor. For the images in Subplots A1-A3 of Fig.~\ref{F10}, their corresponding segmentation results obtained using the proposed and reference algorithms are shown in Subplots C1-C3 and Subplots D1-D3, respectively. Despite the apparent similarity of these segmentations, a closer inspection reveals that the proposed algorithm provides results which better match the expert delineation as compared with the results computed by the reference method. This observation is further supported by the figures in Table~\ref{T3}, which compares the algorithms in terms of the metrics of Section~\ref{metr}.

\begin{table}[htbp]
\caption{Segmentation results pertaining to Section~\ref{occl}}
\centering
\begin{tabular}{ l | l l }
\toprule[1.5pt]
Method	& Weak Priors $\pm \sigma$ & Strong Priors $\pm \sigma$ \\
\midrule
MD 		& $4.60\pm 1.96$ 				& ${\bf 1.55} \pm 2.70$ \\
MAD 	& ${\bf 5.39}\pm 1.51$			& $6.53\pm 2.10$ \\
MAXD	& ${\bf 7.12}\pm 0.708$       		& $9.87\pm 0.849$ \\
SEN		& ${\bf 0.939}\pm 0.032$			& $0.902\pm 0.039$ \\
ACC 	& ${\bf 0.873}\pm 0.029$ 			& $0.850\pm 0.052$ \\
AO 		& ${\bf 0.881}\pm 0.027$   		& $0.858\pm 0.047$ \\
MSE	& ${\bf 0.031} \pm 0.008$ 			& $0.0527\pm 0.019$ \\
\bottomrule
\end{tabular}
\label{T3}
\end{table}

\section{Discussion and Conclusion}
The image segmentation algorithm proposed in this paper can be seen as an extension of the distribution tracking approach of \cite{Freedman04}, similarly to which the proposed method performs segmentation via minimizing a distance between the empirical and model pdf's of image features. Learning the model densities is based on training examples, which consist of a set of representative images and their corresponding segmentation masks. In addition to their standard use as identifiers of the object class, the masks can be also used to estimate the pdf of geometric parameters of the object boundaries. This allows the approach of \cite{Freedman04} to be extended to tracking the ``morphological distributions", in which case the optimal active contour is required to minimize a distance between the model pdf of the geometric parameter(s) and its empirical counterpart. In the current paper, the Bhattacharyya coefficient has been employed to assess the distances between the model and empirical distributions for both photometric and morphological features. Needless to say, the chosen distance can be replaced by other metrics (such as the Kullback-Leibler divergence or the Fisher discriminant \cite{Goudail04}), should there be a rationale behind such a modification.

It goes without saying that, given a pdf of a geometric parameter (such as, e.g., curvature), there exists a myriad of possible shapes whose boundaries will have their empirical pdf's identical to the one above. For this reason, the proposed approach to shape modeling has been coined as ``weak" -- the term which is meant to accentuate the minimally restrictive nature of corresponding shape priors. It has been proven via a series of numerical experiments that the ``weak" modeling is nevertheless capable of effectively regularizing the convergence of active contour under conditions of cluttered and occluded object boundaries. Moreover, as opposed to the PCA-based modeling \cite{Leventon00, Tsai03, Freedman05}, learning the ``weak" priors can be performed using relatively small training sets without noticeably compromising the performance of resulting segmentation. This makes the proposed method particularly useful in situations when one does not have access to an abundance of training shapes.

In practice, the images comprising a given training set are normally registered to a common position, orientation, and scale before a specific shape model can be learned. Within an actual image, on the other hand, the object to be segmented can, in general, appear to be unaligned with respect to such model. For this reason, many segmentation algorithms are bound to use an image registration procedure as a means to bring an evolving active contour into correspondence with its model before the discrepancy between them can be assessed. Needless to say, the registration can substantially increase the overall cost of image segmentation, which is clearly a disadvantage. The ``weak" shape modeling, on the other hand, is free of the above limitation. Indeed, by choosing the shape descriptor to be invariant under a group of geometric transformations, one can effectively forgo the image alignment stage (as long as the latter is  based on the assumed type of transformation). Note that, in the present work, curvature has been used as a shape parameter in the ``weak" modeling, which makes the resulting segmentation invariant under the group of Euclidean transformations. Obviously, the curvature would have to be replaced by a different shape descriptor, should one require the invariance under a different (larger) class of geometric transformations \cite{Sapiro93, Bruckstein97}.  

\bibliographystyle{IEEEtran}
\bibliography{IEEEabrv,references}

\end{document}